%% file: template.tex
\documentclass{article}

\usepackage{arxiv}

\usepackage[utf8]{inputenc} 
\usepackage[T1]{fontenc}    
\usepackage{hyperref}       
\usepackage{url}            
\usepackage{booktabs}       
\usepackage{amsfonts}       
\usepackage{nicefrac}       
\usepackage{microtype}      
\usepackage{lipsum}		
\usepackage{graphicx}
\usepackage[
    backend=biber,
    bibstyle=ieee,
    citestyle=numeric-comp,
    hyperref=true,
    date=year,
    bibencoding=inputenc,
    uniquename=init,
    giveninits=true
]{biblatex}
\usepackage{doi}

\usepackage{mathtools}
\usepackage{subcaption}
\usepackage{accents}
\usepackage{array}
\usepackage{acro}
\usepackage{adjustbox}
\usepackage{siunitx}
\usepackage{changepage}
\usepackage{pgfplots}
\usepackage{pgf-pie}
\usepgfplotslibrary{dateplot}
\usepgfplotslibrary{fillbetween}
\pgfplotsset{compat=newest}
\usetikzlibrary{external}
\acsetup{
	first-style=long-short,
	list/sort=true,
	make-links=false,
	list/template=longtable,
	list/heading=chapter*
}
\DeclareAcronym{PV}{
    short = PV,
    long  = PhotoVoltaic,
    tag = acronyms
}
\DeclareAcronym{CASH}{
    short = CASH,
    long  = Combined Algorithm Selection and Hyperparameter optimization,
    tag = acronyms
}
\DeclareAcronym{MSE}{
    short = MSE,
    long  = Mean Squared Error,
    tag = acronyms
}
\DeclareAcronym{MAE}{
    short = MAE,
    long  = Mean Absolute Error,
    tag = acronyms
}
\DeclareAcronym{nMAE}{
    short = nMAE,
    long  = normalized Mean Absolute Error,
    tag = acronyms
}
\DeclareAcronym{MLP}{
    short = MLP,
    long  = Multi-Layer Perceptron,
    tag = acronyms
}
\DeclareAcronym{GBM}{
    short = GBM,
    long  = Gradient Boosting Machine,
    tag = acronyms
}
\DeclareAcronym{RF}{
    short = RF,
    long  = Random Forest,
    tag = acronyms
}
\DeclareAcronym{SVR}{
    short = SVR,
    long  = Support Vector Regression,
    tag = acronyms
}
\DeclareAcronym{ReLU}{
    short = ReLU,
    long  = Rectified Linear Unit,
    tag = acronyms
}
\DeclareAcronym{TanH}{
    short = TanH,
    long  = Tangent Hyperbolic,
    tag = acronyms
}

\title{AutoPV: Automated photovoltaic forecasts with limited information using an ensemble of pre-trained models}

\author{
    \href{https://orcid.org/0000-0002-9320-5341}{\includegraphics[scale=0.06]{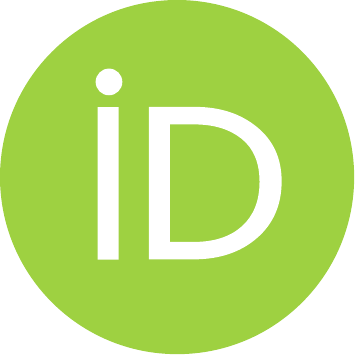}\hspace{1mm}
    Stefan Meisenbacher}\\
	Institute for Automation and Applied Informatics\\
	Karlsruhe Institute of Technology\\
	Eggenstein-Leopoldshafen, 76344, Germany\\
	\texttt{stefan.meisenbacher@kit.edu}\\
	\And
     \href{https://orcid.org/0000-0002-1923-0848}{\includegraphics[scale=0.06]{orcid.pdf}\hspace{1mm}
    Benedikt Heidrich}\\
	Institute for Automation and Applied Informatics\\
	Karlsruhe Institute of Technology\\
	Eggenstein-Leopoldshafen, 76344, Germany\\
	\And
    Tim Martin\\
	Institute for Automation and Applied Informatics\\
	Karlsruhe Institute of Technology\\
	Eggenstein-Leopoldshafen, 76344, Germany\\
	\And
    \href{https://orcid.org/0000-0001-9100-5496}{\includegraphics[scale=0.06]{orcid.pdf}\hspace{1mm}
    Ralf Mikut}\\
	Institute for Automation and Applied Informatics\\
	Karlsruhe Institute of Technology\\
	Eggenstein-Leopoldshafen, 76344, Germany\\
	\And
    \href{https://orcid.org/0000-0002-3572-9083}{\includegraphics[scale=0.06]{orcid.pdf}\hspace{1mm}
    Veit Hagenmeyer}\\
	Institute for Automation and Applied Informatics\\
	Karlsruhe Institute of Technology\\
	Eggenstein-Leopoldshafen, 76344, Germany\\
}

\date{}

\renewcommand{\headeright}{Meisenbacher \textit{et al.}}
\renewcommand{\undertitle}{}
\renewcommand{\shorttitle}{AutoPV: Automated PV forecasts with limited information using an ensemble \phantom{aaaaaaaaaaaaaaaaaaaaaaaaaaaaaaaaaaaa}}

\hypersetup{
    pdftitle={AutoPV: Automated photovoltaic forecasts with limited information using an ensemble of pre-trained models},
    pdfsubject={automated time series forecasting for photovoltaic power generation},
    pdfauthor={Meisenbacher et al.},
    pdfkeywords={
        photovoltaic forecasting,
        ensemble,
        cold-start,
        automated design,
        online operation
    },
    hidelinks
}

\bibliography{references}

\begin{document}
\maketitle

\begin{abstract}
    Accurate \acf{PV} power generation forecasting is vital for the efficient operation of Smart Grids.
    The automated design of such accurate forecasting models for individual \acs{PV} plants includes two challenges:
    First, information about the \acs{PV} mounting configuration (i.e. inclination and azimuth angles) is often missing.
    Second, for new \acs{PV} plants, the amount of historical data available to train a forecasting model is limited (cold-start problem).
    We address these two challenges by proposing a new method for day-ahead \acs{PV} power generation forecasts.
    The proposed AutoPV method is a weighted ensemble of forecasting models that represent different \acs{PV} mounting configurations.
    This representation is achieved by pre-training each forecasting model on a separate \acs{PV} plant and by scaling the model's output with the peak power rating of the corresponding \acs{PV} plant.
    To tackle the cold-start problem, we initially weight each forecasting model in the ensemble equally.
    To tackle the problem of missing information about the \acs{PV} mounting configuration, we use new data that become available during operation to adapt the ensemble weights to minimize the forecasting error.
    The proposed AutoPV method is advantageous as the unknown \acs{PV} mounting configuration is implicitly reflected in the ensemble weights, and only the \acs{PV} plant's peak power rating is required to re-scale the ensemble's output.
    The ensemble approach also allows to represent \acs{PV} plants with panels distributed on different roofs with varying alignments, as these mounting configurations can be reflected proportionally in the weighting.
    Additionally, the required computing memory is decoupled when scaling AutoPV to several hundreds of \acs{PV} plants, which is beneficial in Smart Grid environments with limited computing capabilities.
    For a real-world data set with 11 \acs{PV} plants, the accuracy of AutoPV is comparable to an individual model trained on two years of data and outperforms an incrementally trained individual model.
\end{abstract}

\keywords{photovoltaic forecasting \and ensemble \and cold-start \and automated design \and online operation}
\twocolumn
\acresetall

\section{Introduction}
\label{sec:introduction}

The Paris Climate Agreement targets net-zero emissions of the energy system by 2050 \cite{UNFCCC2015}.
This target requires reducing primary energy demand, increasing energy efficiency, and increasing the share of renewables in electricity generation.
The latter endangers grid stability, as the fluctuating and non-controllable production behavior of renewables may put the frequency stability under pressure or lead to power line congestion.
Hence, a Smart Grid is required that integrates communication, measurement, control, and automation technology.
This integration enables the Smart Grid to monitor local power grid conditions in real-time and enables full utilization of the grid capacity by balancing demand and supply \cite{SmartGrid}.
The full utilization and balance of supply and demand are realizable by using demand-side management and proactive supply control.
Yet, both rely on accurate demand and supply forecasts \cite{Dannecker2015}.

In this context, the automated design of such accurate forecasting models for the decentral generation of \ac{PV} power includes two challenges related to missing information about an individual \ac{PV} plant:
First, the \ac{PV} mounting configuration characterized by inclination and azimuth angles is difficult to obtain as both are often incompletely documented or roughly approximated.
Especially for \ac{PV} plants with panels distributed on different roofs with varying alignments, often only the total peak power rating of the \ac{PV} plant is known.
This missing documentation limits the large-scale application of forecasting methods that rely on information about the \ac{PV} mounting configuration (e.g. \cite{Perpinan2012, Holmgren2018}).
Second, often historical data for training the forecasting model are not available (cold-start problem).
For example, for newly built \ac{PV} plants, nearly no data are available.
Consequently, data-intensive forecasting methods such as deep learning methods (e.g. \cite{AbdelBasset2021, Aslam2021, Wang2019}) are not directly applicable to new \ac{PV} plants.

Therefore, the present paper addresses these two challenges by proposing a new method for day-ahead \ac{PV} forecasts called AutoPV.
This method only requires the new \ac{PV} plant’s peak power rating, is applicable without any training data of the new plant, and adapts itself during operation.
Our proposed forecasting method AutoPV is based on creating an ensemble of multiple forecasting models.
The models in the ensemble pool are pre-trained on historical data of different \ac{PV} plants of the same region with various mounting configurations that are scaled with the corresponding peak power rating.
Thus, the model pool contains expert models for various \ac{PV} mounting configurations.
To tackle the cold-start problem where no data of the \ac{PV} plant are available, each model in the ensemble contributes equally to the forecast.
During the operation, new data are collected and used to adapt the ensemble weights.
More specifically, the contribution of each model in the ensemble pool is adapted such that the weighted sum optimally fits the new \ac{PV} plant.
Thereby, the unknown \ac{PV} mounting configuration is implicitly reflected in the ensemble weights.
Also, \ac{PV} plants with panels distributed on different roofs with varying alignments can be reflected proportionally in the weighting.

Regarding the taxonomy of automated forecasting pipelines described in \cite{AutomatedForecastingPipeline}, our method covers all 5 sections, i.e., data pre-processing, feature engineering, hyperparameter optimization, forecasting method selection, and forecast ensembling.
The organization of this workflow using a machine learning pipeline, as well as automatic model design and selection along with continuous adaptation during operation, corresponds to Automation level 3 \cite{AutomationLevels}.
For creating automated forecasting pipelines, generic frameworks such as TPOT \cite{TPOT} exist, but they are not adapted to a specific domain and thus cannot leverage prior knowledge on \ac{PV} forecasting.

The remainder of this paper is organized as follows.
We describe the methodology of the proposed \ac{PV} power generation forecasting method in \autoref{sec:methodology}, evaluate it in \autoref{sec:evaluation}, discuss the results in \autoref{sec:discussion} and give a conclusion and an outlook in \autoref{sec:conclusion_outlook}.

\section{Design of the AutoPV method}
\label{sec:methodology}
As described in the introduction, unknown \ac{PV} mounting configurations characterized by inclination and azimuth angles are challenging for the design of a \ac{PV} forecasting model.
Thus, the underlying idea of the proposed AutoPV method is that each new \ac{PV} mounting configuration can be described by the sum of weighted elements from a sufficiently diverse pool of forecasting models of the same region.

The proposed AutoPV method incorporates three steps: i) create the model pool, ii) form the ensemble forecast by an optimally weighted sum of the scaled forecasts, and iii) re-scale the ensemble forecast with the new \ac{PV} plant's peak power rating.
These three steps are detailed in the following and \autoref{fig:ensemble_method} provides exemplary results of these three steps.

\subsection{Creation of the ensemble pool models}
Each model in the ensemble pool is equally designed.
However, each model is pre-trained on historical data of an individual \ac{PV} plant.
Thus, the ensemble model pool covers a diverse set of \ac{PV} mounting configurations.
Each model is based on a machine learning pipeline, including pre-processing and feature extraction, an automatically designed regression estimator, and a set of rules reflecting prior knowledge.
First, we describe the machine learning pipeline and afterward the automated regression estimator design.

\paragraph{Machine learning pipeline}

For each model $n\ldots N$, we pre-process the training data by scaling the \ac{PV} power generation measurement according to the peak power rating $P_{\text{n},n}$ of the corresponding \ac{PV} plant
\begin{equation}
    \underaccent{\bar}{y}_n[k] =  \frac{y_n[k]}{P_{\text{n},n}},
    \label{eq:scaling}
\end{equation}
where $k \in \mathbb{N}^K$ is the sample index.
In contrast to the \ac{PV} mounting configurations, we assume that $P_{\text{n},n}$ is a parameter of the \ac{PV} plants that can be reliably determined.
However, scaling with $P_{\text{n},n}$ does not lead to aligned \ac{PV} power generation curves (see \autoref{fig:ensemble_pool_model_forecasts_normalized}).
This is because the mounting configuration of a \ac{PV} plant influences the maximum possible amount of generated power.
For example, a \ac{PV} plant oriented to the south generates more than a \ac{PV} plant oriented to the west.
In addition, global radiation's seasonal and weather-dependent intensity also affects the distance between the \ac{PV} power output and the plant's peak power rating.

As inputs, each regression estimator uses global radiation $\hat{G}$ considering cloud cover and air temperature $\hat{T}$ forecasts, and the corresponding second-order polynomial and interaction features.
Additionally, we use sin-cos encoded cyclic features of the month and minute of the day to represent seasonal information:
\begin{flalign}
    x_\text{s12}[k]     &= \sin (2 \pi \cdot \text{month}[k] \cdot 12^{-1}),\\
    x_\text{c12}[k]     &= \cos (2 \pi \cdot \text{month}[k] \cdot 12^{-1}),\\
    x_\text{s1440}[k]   &= \sin (2 \pi \cdot \text{minute}[k] \cdot 1440^{-1}),\\
    x_\text{c1440}[k]   &= \cos (2 \pi \cdot \text{minute}[k] \cdot 1440^{-1}).
\end{flalign}
The periodicity of these trigonometric functions establishes similarities between temporally related samples, while the sin-cos pair is necessary because, otherwise, the encoding would be ambiguous at several points.\footnote{
    We do not consider historical data as lag features.
    The reason is that the models in the ensemble pool are pre-trained on data of different \ac{PV} plants.
    Consequently, using historical data would require getting data from all used \ac{PV} plants in operation.
}

The resulting regression estimator for plant $n$ in the ensemble pool is a function of
\begin{equation}
\begin{split}
    \hat{\underaccent{\bar}{y}}_n[k] = f
    &\left(
        \hat{G}^2[k],
        \hat{G}[k],
        (\hat{G} \cdot \hat{T})[k],
        \hat{T}[k],
        \hat{T}^2[k],
    \right.\\
    &\left.\vphantom{
            \hat{G}^2[k], \hat{G}[k],
            (\hat{G} \cdot \hat{T})[k],
            \hat{T}[k],
            \hat{T}^2[k]}
        x_\text{s12}[k],
        x_\text{c12}[k],
        x_\text{s1440}[k],
        x_\text{c1440}[k],
        \boldsymbol{p}\right),
    \label{eq:model_structure}
\end{split}
\end{equation}
where the parameters $\boldsymbol{p}$  are determined by training the machine learning pipeline on historical data.

To consider prior knowledge in forecasting \ac{PV} power generation, we use two rules.
First, no \ac{PV} power is generated if there is no solar radiation.
Second, negative \ac{PV} power generation is not possible.
Consequently, we drop the night times from the training data and set the forecast to zero during these times. Furthermore, the negative values in the forecast are set to zero.\footnote{
    Slightly negative values may occur in the first data points at sunrise and sunset.
}

Incorporating these two steps leads to
\begin{equation}
    \hat{\underaccent{\tilde}{y}}_n[k] =
        \begin{cases*}
            \hat{\underaccent{\bar}{y}}_n[k],  & if $\hat{G}[k] > 0$ and $\hat{\underaccent{\bar}{y}}_n[k] > 0$,\\
            0           & otherwise,\\
        \end{cases*}
    \label{eq:model_structure_corrected}
\end{equation}
as the final step in the machine learning pipeline.

\paragraph{Automated regression estimator design}

For automatically designing the regression estimator, we define a \ac{CASH} problem.
In the \ac{CASH} problem, we aim to minimize the machine learning pipeline's \ac{MSE}
\begin{equation}
    \text{MSE} = \frac{1}{K} \sum_{k=1}^K\left(\hat{\underaccent{\tilde}{y}}_n[k] - \underaccent{\bar}{y}_n[k]\right)^2,
    \label{eq:mean_squared_error}
\end{equation}
by selecting the optimal configuration, i.e., the optimal regression algorithm and the corresponding optimal hyperparameters.
The configuration space is shown in \autoref{tab:configuration_space}, which is explored during \ac{CASH} using Bayesian optimization.
Each configuration trial is trained on a training data set and assessed on a hold-out validation data set.

\paragraph{Implementation}

The machine learning pipeline is implemented using the Python package pyWATTS \cite{pyWATTS} and the regression estimators are implemented using the Python package scikit-learn \cite{sklearn}.
For solving the \ac{CASH} problem, we use the Python package Ray Tune \cite{ray-tune} with the hyperopt \cite{hyperopt} search algorithm.
The \ac{CASH} is automatically stopped if the \ac{MSE} plateaus across trials, i.e., the \ac{MSE} of the top 10 trials have a standard deviation of less than 0.001 with a patience of 15 trials.

\begin{table}
	\caption{The configuration space for the automated regression estimator design using \acf{CASH} in scikit-learn \cite{sklearn} naming convention.}
    \label{tab:configuration_space}
	\begin{adjustbox}{max width=\linewidth}
    	\input{tables/configuration_space}
	\end{adjustbox}
	\tiny\\[0.5mm]
\end{table}

\subsection{Optimal weighting of the ensemble pool models}

The idea of creating an ensemble is to increase the robustness of data-driven models by combining multiple forecasts from a pool of different models \cite{Shaub2020}.
Apart from weighting each model in the pool equally (averaging), one may give more weight to models from which we expect good performance and give less weight to models from which we expect poor performance, i.e., 
\begin{equation}
    \hat{\underaccent{\tilde}{y}}[k] = \sum^N_{n=1} w_n \cdot \hat{\underaccent{\tilde}{y}}_n[k]
    \label{eq:weighted_sum}
\end{equation}
with $w_n$ being the weight of the $n$-th model in the ensemble pool.

To find the weights of the ensemble, we distinguish the initialization and the operation phase.
In the initialization phase, historical power generation data for the new \ac{PV} plant are not available.
Thus, we weight each model in the ensemble pool equally.
During the operation, new data are used to cyclically adapt the ensemble weights in such a way that they optimally fit the new \ac{PV} plant (see \autoref{fig:ensemble_forecast_normalized}).
For optimal weighting, we vary the weights of the ensemble pool models $\boldsymbol{w}$ to minimize the \ac{MSE} over the $K$ most recent samples of the $P_\text{n}$-scaled data $\underaccent{\tilde}{\boldsymbol{y}}$.
With \eqref{eq:weighted_sum}, this results in the optimization problem
\begin{equation}
\begin{split}
    \min_{\boldsymbol{w}} \frac{1}{K} \sum^K_{k=1} \left(\hat{\underaccent{\tilde}{y}}[k] - \underaccent{\bar}{y}[k] \right)^2
    \\
    \textrm{s.t.} \quad \boldsymbol{w} \in [0,1], \sum^N_{n=1} w_n = 1,
    \label{eq:ensemble_optimization}
\end{split}
\end{equation}
which we solve using the least squares implementation of the Python package SciPy \cite{scipy}, and normalize the weights afterward to hold the constraints.

Note that the cycle length $C$ of the adaption routine and the number of considered samples $K$ are hyperparameters of the proposed AutoPV.

\subsection{Transformation of the ensemble forecast to the new \ac{PV} plant}
\begin{figure}[ht!]
        \begin{subfigure}[b]{0.5\textwidth}
            \includegraphics[scale=1]{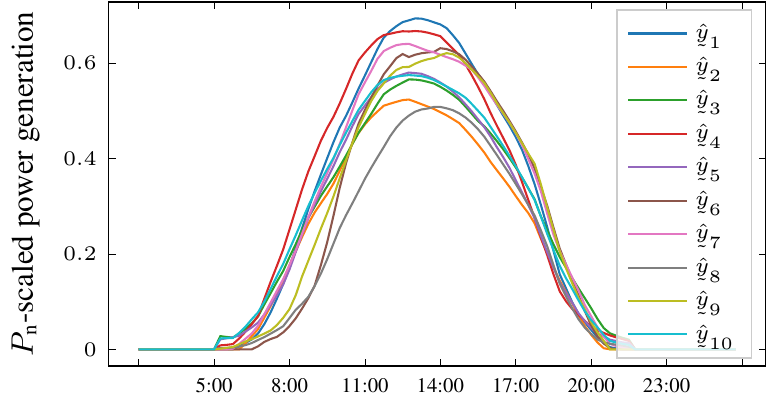}
            \caption{The result of the first step is the scaled ensemble pool model forecasts $\hat{\underaccent{\tilde}{y}}_n$.}
            \label{fig:ensemble_pool_model_forecasts_normalized}
        \end{subfigure}%
        \\[2mm]
        \begin{subfigure}[b]{0.5\textwidth}
            \includegraphics[scale=1]{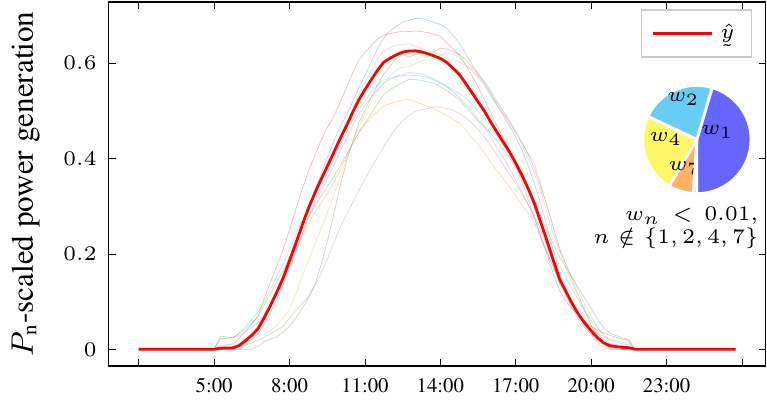}
            \caption{The result of the second step is the scaled ensemble forecast $\hat{\underaccent{\tilde}{y}}$.}
            \label{fig:ensemble_forecast_normalized}
        \end{subfigure}%
        \\[2mm]
        \begin{subfigure}[b]{0.5\textwidth}
            \includegraphics[scale=1]{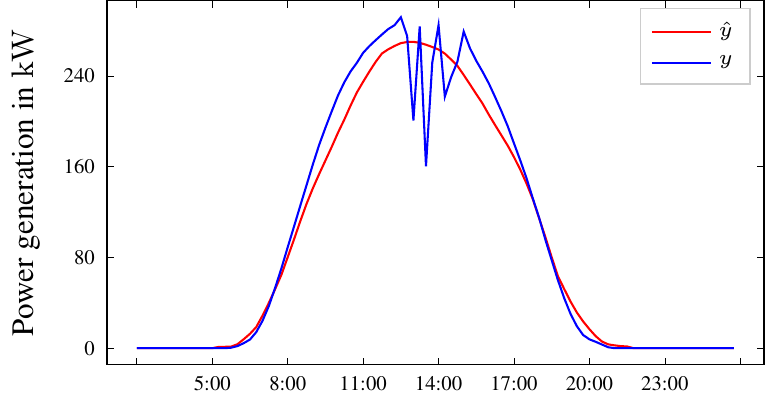}
            \caption{The result of the third step is the re-scaled ensemble forecast $\hat{{y}}$. Additionally, this plot shows the measurement $y$ with cloud overflight between 12:00 and 15:00. The ensemble forecast is based on a day-ahead radiation forecast that also considers cloud cover but with lower spatial and temporal resolution.}
            \label{fig:ensemble_forecast}
        \end{subfigure}
    \caption{The three steps of the proposed \ac{PV} ensemble forecasting method exemplified for \ac{PV} plant no. 11.}
    \label{fig:ensemble_method}
\end{figure}

The last step yields the final ensemble forecast, shown in \autoref{fig:ensemble_forecast}).
To this end, it re-scales the weighted average of the ensemble according to the peak power rating $P_\text{n,new}$ of the new \ac{PV} plant:
\begin{equation}
    \hat{y}[k] = \hat{\underaccent{\tilde}{y}}[k] \cdot P_\text{n,new}.
    \label{eq:re_scaling}
\end{equation}

\section{Evaluation}
\label{sec:evaluation}

We evaluate the proposed AutoPV method on a real-world data set containing three years (2018 -- 2020) of quarter-hourly power generation measurement ($\si{\kWh}$) from 11 \ac{PV} plants located in southern Germany.
For better interpretability, we transform the energy generation time series into the mean power generation time series
\begin{equation}
    \overline{P}[k] = \frac{\Delta E[k]}{t_k},
\end{equation}
where $\Delta E[k]$ is the energy generated in the sample period $t_k$ ($1/\si{\hour}$).
Additionally, we use corresponding day-ahead weather forecasting data.

We pre-train the ensemble pool models with the data from 2018 and 2019, and assess the proposed AutoPV method with all samples of 2020.
In pre-training, the automated regression estimator design via \ac{CASH} predominately creates a \ac{MLP} regressor with two or three hidden layers and the \ac{ReLU} activation function.

In the following, we first evaluate the performance of AutoPV and then the consistency of AutoPV's ensemble optimization.

\paragraph{Performance evaluation}

We evaluate the performance of AutoPV using a plant-wise leave-one-out evaluation. 
That is, we assess the forecasting error of the proposed AutoPV method on each \ac{PV} plant while excluding this plant from the ensemble model pool.
Since the \ac{PV} plants in the data set have different peak power ratings and would differently influence the MAE, we use the \ac{nMAE} as assessment metric:
\begin{equation}
    \text{nMAE} = \frac{\sum_{k=1}^K\left|\hat{y}[k]-y[k]\right|}{\sum_{k=1}^K{y[k]}}.
    \label{eq:normalized_mean_absolute_error}
\end{equation}

In the evaluation, we start with equal ensemble weights and adapt them every 28 days based on the samples obtained in the most recent 28 days ($C=28, K=28 \cdot 96$).

We compare the performance of the proposed AutoPV method with three methods.
The first method reflects an ideal situation in the sense that we assume that there are two years of historical data available (2018, 2019) to train an individual model (IM-HDA) for the considered \ac{PV} plant.
The second method is an averaging ensemble, which reflects the initialization phase of AutoPV without adapting the weights during operation.
As a third method, we train an individual model incrementally (IM-IT) every 28 days.
More specifically, the IM-IT uses all data of the testing data set (2020) that is obtained until the $l$-th adaption ($C=28, K=l \cdot 28 \cdot 96, l=[1,2,\ldots]$).
The performance comparison of IM-IT, Averaging, and AutoPV with the IM-HDA is unfair in terms of training data because the IM-HDA is trained on data of the \ac{PV} plant that is unavailable to them.

\begin{figure}[b!]
    \includegraphics[scale=1]{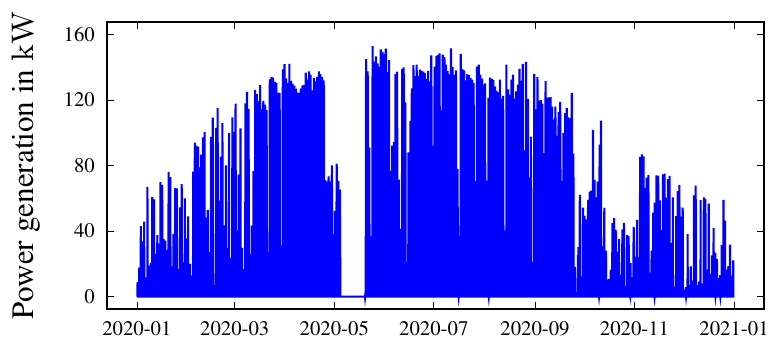}
    \caption{Shutdowns and partial shutdowns of \ac{PV} plant no. 7 lead to sudden dips in the generated power.}
    \label{fig:original_pv6}
\end{figure}

In the performance evaluation, we make three observations.
First, considering the average assessment metric over all plants $\overline{\text{nMAE}}$, we see that the proposed AutoPV method comes close to the ideal situation IM-HDA.
Second, regarding the cold-start problem, the proposed AutoPV method outperforms the IM-IT (compare \autoref{tab:results}).
Third, \ac{PV} plant no. 7 has multiple dips in September and October, as well as a complete shutdown of this plant in May (see \autoref{fig:original_pv6}), and the \ac{nMAE} of all methods is comparatively high for this plant.

\begin{table}
	\caption{The \acf{nMAE} of the proposed AutoPV method compared to the historical data available model (IM-HDA), the averaging ensemble, and the incrementally trained individual model (IM-IT) in the plant-wise leave-one-out evaluation.}
	\label{tab:results}
	\begin{adjustbox}{max width=\linewidth}
	\input{tables/results_leave-one-out}
    \end{adjustbox}
    \scriptsize\\[0.5mm]
    \begin{adjustwidth}{11em}{1em}%
    bolt: winner of \textbf{fair} comparison in terms of training data; bolt\&italics: winner of \textit{\textbf{unfair}} comparison
    \end{adjustwidth}
\end{table}

\paragraph{Consistency evaluation}

We evaluate the consistency of the AutoPV ensemble optimization as described above, except that the ensemble model pool includes all \ac{PV} plants, i.e., the IM-HDA is available in the pool.
Hence, a consistent ensemble optimization should give high weight to the IM-HDA.

\begin{figure}[t!]
        \begin{subfigure}[b]{0.5\textwidth}
            \includegraphics[scale=1]{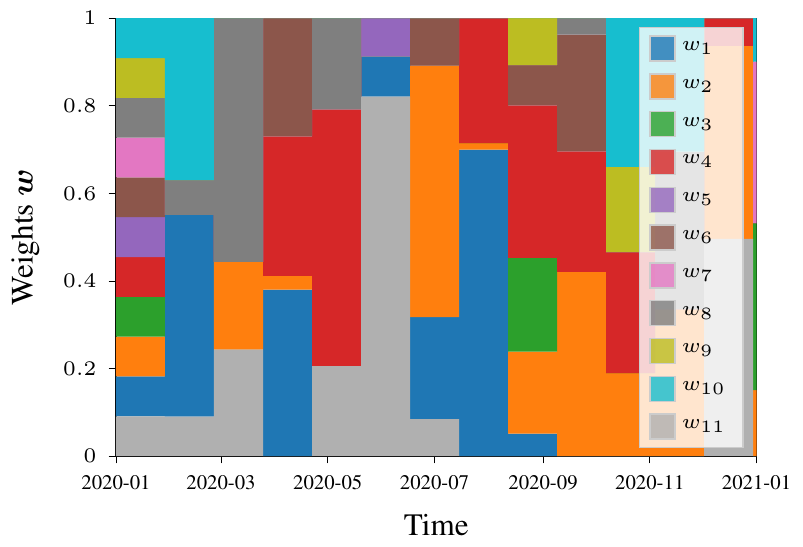}
            \caption{\ac{PV} plant no. 5 is rather similar compared to other plants.}
            \label{fig:weights_similar_plants}
        \end{subfigure}%
        \\[0mm]
        \begin{subfigure}[b]{0.5\textwidth}
            \includegraphics[scale=1]{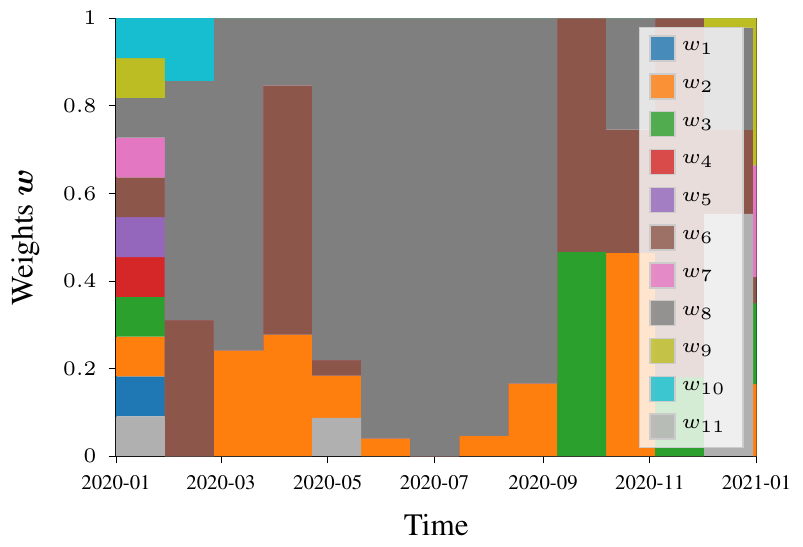}
            \caption{\ac{PV} plant no. 8 is rather unique compared to other plants.}
            \label{fig:weights_unique_plants}
        \end{subfigure}%
    \caption{Weight development in the consistency evaluation.}
    \label{fig:weights_consistency}
\end{figure}

\autoref{fig:weights_consistency} shows the exemplary weight evolution of \ac{PV} plant no. 5 and \ac{PV} plant no. 8.
\ac{PV} plant no. 5 represents a plant whose power generation curve is rather similar to other plants, while \ac{PV} plant no. 8 reflects a rather unique power generation curve (compare in \autoref{fig:ensemble_pool_model_forecasts_normalized}).

In the consistency evaluation, we make three observations.
First, after the initialization phase, the weights change significantly.
Second, the more unique the power generation curve of a \ac{PV} plant, the clearer the weighting towards the IM-HDA (compare \autoref{fig:weights_similar_plants} and \autoref{fig:weights_unique_plants}, where for \ac{PV} plant no. 5 the weight $w_5$ and for \ac{PV} plant no. 8 the weight $w_8$ corresponds to the respective IM-HDA).
Third, the weighting is generally clearer towards the IM-HDA in the summer than in the winter.

\section{Discussion}
\label{sec:discussion}

This section discusses the results, limitations, and benefits of the proposed AutoPV method.

With regard to the results, we discuss two aspects:
First, we observe that the proposed AutoPV method and the ideal IM-HDA perform almost similarly.
Thus, we conclude that any \ac{PV} mounting configuration in the given dataset can be represented by recombining the ensemble pool models' forecasts.
This recombination via the ensemble weights also allows to represent \ac{PV} plants with panels distributed on different roofs with varying alignments.
More specifically, the AutoPV method is not limited to represent a single \ac{PV} mounting configuration.
Instead, the mounting configurations of \ac{PV} panels distributed on different roofs can be reflected proportionally in the weighting.
Second, we observe that the proposed AutoPV method outperforms the IM-IT.
This might be explained by the fact that the pre-trained ensemble pool models already reflect the entire seasonality, whereas the IM-IT learns it with a delay.
However, the forecasting accuracy of both is harmed by dips in the generated \ac{PV} power, as this sudden change is only captured with a delay by the cyclic adaptation.

Regarding the limitations, we discuss the creation of the ensemble pool.
For representing a new \ac{PV} plant by recombining the ensemble pool models' forecasts, they must be diverse with respect to the \ac{PV} mounting configurations. 
To create a diverse ensemble model pool, we can use prior knowledge or select a diverse set of curves after the $P_\text{n}$-scaling (see \autoref{fig:ensemble_pool_model_forecasts_normalized}).
However, we cannot guarantee that an ensemble pool of 10 models is sufficiently diverse to represent all alignment combinations.
This limitation can be overcome by automatically designing an individual model for the new \ac{PV} plant after a full year and adding it to the ensemble pool.

The proposed AutoPV method is beneficial because it is universally applicable and achieves promising results with only 10 representative models in the ensemble pool.
As scaling AutoPV to several hundred \ac{PV} plants still requires only the 10 ensemble pool models to be stored in the computing memory, and the weight adaptation effort is very low, AutoPV is advantageous for Smart Grid environments with limited computing capabilities.
Moreover, the universality of our approach allows us to use arbitrary complex models in the ensemble pool.

\section{Conclusion and outlook}
\label{sec:conclusion_outlook}
Designing accurate \ac{PV} power generation forecasting models for Smart Grid operation includes two challenges:
The first is the missing information about the \ac{PV} mounting configuration (inclination and azimuth angles).
The second is the limited availability of historical data to train the forecasting model (cold-start problem).
To address both challenges, we propose a new method for day-ahead \ac{PV} forecasts that only requires the new \ac{PV} plant’s peak power rating, that is applicable without any training data of the new plant, and that adapts itself during operation (Automation level 3 \cite{AutomationLevels}).
The new AutoPV method is an automated forecasting pipeline \cite{AutomatedForecastingPipeline} based on an ensemble with a pool of pre-trained models that represent various \ac{PV} mounting configurations.
While initially, each model in the ensemble pool contributes equally to the forecast, the contribution of each model is adapted during operation using the least squares method such that the weighted sum optimally fits the new \ac{PV} plant.
Hence, the AutoPV method is also applicable to \ac{PV} plants with panels distributed on different roofs with varying alignments, as these mounting configurations can be reflected proportionally in the weighting.

The evaluation on real-world data shows that the proposed AutoPV method outperforms an incrementally trained individual model.
It achieves comparable performance to the ideal situation of having two years of historical data available to train an individual model.
The evaluation also shows that for reliable and automated online operation, it is necessary to consider \ac{PV} power generation dips caused by technical defects, scheduled maintenance, or soiling.
Therefore, the AutoPV method will be extended in future research to include a drift detection method and consider the \ac{PV} plant's efficiency with additional weight in the ensemble optimization.

\section*{Acknowledgments}
This project is funded by the Helmholtz Association’s Initiative and Networking Fund through Helmholtz AI the Helmholtz Association under the Program ``Energy System Design''.
Furthermore, the authors thank Stadtwerke Karlsruhe Netzservice GmbH (Karlsruhe, Germany) for the data required for this work. 

\printbibliography

\end{document}

%% file: tables/configuration_space.tex
\begin{tabular}{rll}
\toprule
\multicolumn{1}{l}{\textbf{Regression estimator}} & \textbf{Hyperparameter} & \textbf{Value range} \\
\midrule
\multicolumn{1}{l}{\texttt{Ridge}} & \texttt{alpha} & $[0.05, 1]$ \\
\multicolumn{1}{l}{\texttt{MLPRegressor}} & \texttt{activation} & $\{\texttt{logistic, tanh, relu}\}$ \\
      & \texttt{hidden\_layer\_sizes} & $\{([10, 100])$ \\
      &       & $\phantom{\{}([10, 100], [10, 100])$ \\
      &       & $\phantom{\{}([10, 100], [10, 100], [10, 100])\}$ \\
\multicolumn{1}{l}{\texttt{GradientBoostingRegressor}} & \texttt{learning\_rate} & $[0.01, 1]$ \\
      & \texttt{n\_estimators} & $[10, 300]$ \\
      & \texttt{max\_depth} & $[1, 10]$ \\
\multicolumn{1}{l}{\texttt{RandomForestRegressor}} & \texttt{n\_estimators} & $[10, 300]$ \\
      & \texttt{max\_depth} & $[1, 10]$ \\
\multicolumn{1}{l}{\texttt{SVR}} & \texttt{C} & $[0.01, 10]$ \\
      & \texttt{epsilon} & $[0.001, 1]$ \\
\bottomrule
\end{tabular}%

%% file: tables/results_leave-one-out.tex
\begin{tabular}{rc|ccc}
\cmidrule{2-5}      & \textbf{IM-HDA} & \textbf{IM-IT} & \textbf{Averaging} & \textbf{AutoPV} \\
\cmidrule{2-5} Training data & 2 years & 1 month & none & none \\
Adaption cycle $C$ & none & 28d & none & 28d \\
Adaption samples $K$ & none & $l \cdot 28 \cdot 96$ & none & $28 \cdot 96$ \\
\cmidrule{2-5}
$\text{nMAE}_{1\phantom{1}}$ & 0.308 & 0.332 & 0.367 & \textbf{0.278} \\
$\text{nMAE}_{2\phantom{1}}$ & \textit{\textbf{0.298}} & 0.321 & 0.302 & \textbf{0.299} \\
$\text{nMAE}_{3\phantom{1}}$ & \textit{\textbf{0.271}} & 0.310 & 0.314 & \textbf{0.289} \\
$\text{nMAE}_{4\phantom{1}}$ & 0.316 & 0.349 & 0.385 & \textbf{0.291} \\
$\text{nMAE}_{5\phantom{1}}$ & \textit{\textbf{0.325}} & 0.343 & 0.353 & \textbf{0.335} \\
$\text{nMAE}_{6\phantom{1}}$ & 0.327 & 0.322 & 0.299 & \textbf{0.289} \\
$\text{nMAE}_{7\phantom{1}}$ & \textit{\textbf{0.418}} & 0.588 & 0.491 & \textbf{0.448} \\
$\text{nMAE}_{8\phantom{1}}$ & 0.280 & 0.310 & 0.280 & \textbf{0.277} \\
$\text{nMAE}_{9\phantom{1}}$ & \textit{\textbf{0.362}} & 0.424 & 0.502 & \textbf{0.416} \\
$\text{nMAE}_{10}$ & \textit{\textbf{0.307}} & 0.365 & 0.364 & \textbf{0.318} \\
$\text{nMAE}_{11}$ & \textit{\textbf{0.311}} & 0.471 & 0.324 & \textbf{0.311} \\
\cmidrule{2-5}
$\overline{\text{nMAE}}_{\phantom{1}\phantom{1}}$ & \textit{\textbf{0.320}} & 0.376 & 0.362 & \textbf{0.323} \\
\end{tabular}%